\documentclass[runningheads]{llncs}
\usepackage{booktabs}
\usepackage[T1]{fontenc}
\usepackage{longtable}
\usepackage{hyperref}
\usepackage{nameref}
\usepackage{amsmath}
\usepackage{amssymb}
\usepackage{multirow} 
\usepackage{adjustbox}
\usepackage{algorithm}
\usepackage{framed}

\usepackage{algorithmic}
\usepackage[numbers]{natbib}

\usepackage{graphicx}
\begin{document}
%
\title{MeGA: Merging Multiple Independently Trained Neural Networks Based on Genetic Algorithm}
\titlerunning{Merging Multiple Independently Networks}
%
\author{Daniel Yun\inst{1}}
\authorrunning{Yun}
%
\institute{Stony Brook University, Stony Brook NY 11794, USA\\
\email{juyoung.yun@stonybrook.edu}}
\maketitle              
\begin{abstract}
In this paper, we introduce a novel method for merging the weights of multiple pre-trained neural networks using a genetic algorithm called MeGA. Traditional techniques, such as weight averaging and ensemble methods, often fail to fully harness the capabilities of pre-trained networks. Our approach leverages a genetic algorithm with tournament selection, crossover, and mutation to optimize weight combinations, creating a more effective fusion. This technique allows the merged model to inherit advantageous features from both parent models, resulting in enhanced accuracy and robustness. Through experiments on the CIFAR-10 dataset, we demonstrate that our genetic algorithm-based weight merging method improves test accuracy compared to individual models and conventional methods. This approach provides a scalable solution for integrating multiple pre-trained networks across various deep learning applications. Github is available at: \href{https://github.com/YUNBLAK/MeGA-Merging-Multiple-Independently-Trained-Neural-Networks-Based-on-Genetic-Algorithm}{Here}

\keywords{Neural Networks \and Deep Learning \and Genetic Algorithm \and Evolutionary Algorithms}
\end{abstract}

\section{Introduction}
In recent years, deep learning has achieved state-of-the-art performance across various tasks, such as image classification and natural language processing, largely due to the use of pre-trained neural networks~\cite{Goodfellow2016DeepLearning, lecun2015deep, schmidhuber2015deep}. These networks can be fine-tuned for specific tasks, saving computational resources and time. However, effectively combining multiple pre-trained models of the same architecture to harness their collective strengths and mitigate individual weaknesses remains a significant challenge.

Merging weights from different pre-trained models with identical architectures is crucial because individual models often learn complementary features from the data~\cite{Hansen1990Neural, dietterich2000ensemble, liu2020survey}. Combining these models can create a more robust and accurate system, leveraging the strengths of each model and leading to improved performance. Additionally, model fusion can help achieve better generalization by averaging out individual biases and reducing overfitting~\cite{dietterich2000ensemble, lakshminarayanan2017simple, welling2011bayesian}. 

Training models in parallel rather than sequentially offers efficiency and practicality. Multiple models can learn diverse patterns quickly, and merging their weights can achieve superior performance faster~\cite{huang2017snapshot, dean2012large, goyal2017accurate}. This approach is particularly advantageous in distributed learning environments, facilitating scalable and efficient training across multiple GPUs or devices~\cite{dean2012large, li2014scaling}. 

The primary challenge in weight merging is finding the optimal combination that maximizes performance without further training or altering the model's architecture. Simple averaging fails to account for the intricate dependencies within the neural network~\cite{Izmailov2018Averaging, fort2019stiffness, wang2020towards}. To address this, we propose a novel approach using a genetic algorithm to optimize weight merging. Genetic algorithms efficiently search large, complex spaces by iteratively selecting, combining, and mutating weights~\cite{Mitchell1998AnIntroduction, holland1992adaptation, goldberg1989genetic}. 

In this paper, we introduce our genetic algorithm-based method, MeGA, for merging the weights of multiple trained CNN models. Experiments on the CIFAR-10 dataset~\cite{Krizhevsky2009Learning} demonstrate improvements in test accuracy compared to individual models and traditional techniques. Rather than focusing on comparative analysis, this research emphasizes the methodology of effectively merging models to leverage their collective strengths. Our results highlight the potential of genetic algorithms in optimizing neural network weight fusion, providing a scalable solution for integrating multiple pre-trained models in various deep learning applications. Additionally, we demonstrate that it is possible to merge the weights of neural networks that have been initialized differently and trained independently, further underscoring the flexibility and robustness of our proposed method.

\section{Related Works}
The fusion of neural network models has been an active area of research due to its potential to enhance model performance by leveraging the strengths of individual models. Several methods have been proposed to address the challenges associated with model merging, each with unique approaches and varying degrees of success. \\

\noindent \textbf{Weight Merge.}
One of the earliest and simplest methods for model fusion is weight averaging, where the weights of two or more models are averaged to create a new model. This approach, while straightforward, often fails to consider the complex interactions between different layers of the networks. Goodfellow et al. introduced an approach that averages the weights of neural networks trained with different initializations, showing marginal improvements in performance \citep{goodfellow2014explaining}. Similarly, ensemble methods, where the predictions of multiple models are combined, have been extensively studied and are known to improve model robustness and accuracy \citep{hansen2010neural}. However, these methods do not directly merge the models' internal representations, potentially limiting their effectiveness. Izmailov et al. proposed Stochastic Weight Averaging (SWA), which improves generalization by averaging weights along the trajectory of SGD with a cyclical or constant learning rate \citep{Izmailov2018Averaging}. Welling and Teh applied Stochastic Gradient Langevin Dynamics (SGLD) to sample from the posterior distribution of model weights, thereby integrating Bayesian principles into model merging \citep{welling2011bayesian}. This method helps in capturing the model uncertainty but is computationally intensive and complex to implement. Huang et al. introduced the concept of Snapshot Ensembles, where a single neural network is trained with a cyclical learning rate schedule, and multiple snapshots of the model are taken at different local minima \citep{huang2017snapshot}. Shen and Kong~\cite{shen2004optimizing} demonstrated the use of genetic algorithms to enhance prediction accuracy by optimizing weights in neural network ensembles. \\

\noindent \textbf{Genetic Algorithms in Neural Networks.} Genetic algorithms have been applied in various aspects of neural network optimization. Real et al. utilized evolutionary algorithms for neural architecture search, demonstrating the potential of genetic approaches in discovering optimal network structures \citep{real2017large}. Similarly, Xie and Yuille used genetic algorithms for model pruning, showing that evolutionary techniques can effectively optimize neural network parameters \citep{xie2017genetic}. Stanley and Miikkulainen introduced NeuroEvolution of Augmenting Topologies (NEAT), which evolves both the architecture and weights of neural networks, leading to improved performance on complex tasks \citep{stanley2002evolving}. Another notable application is by Fernando et al., who proposed PathNet, a method that uses evolutionary strategies to discover pathways through a neural network, enabling efficient transfer learning \citep{fernando2017pathnet}. Finally, Loshchilov and Hutter applied a genetic algorithm for hyperparameter optimization, showing that evolutionary strategies can outperform traditional grid and random search methods \citep{loshchilov2016cma}. \\

Building on the success of these methods, our approach leverages genetic algorithms to optimize the weight merging process of pre-trained CNN models. By iteratively selecting, combining, and mutating weights, our method systematically explores the weight space to discover an optimal or near-optimal set of weights. This novel approach not only enhances model performance but also facilitates efficient parallel training and distributed learning across multiple devices, addressing scalability and robustness issues inherent in traditional methods.

\section{Methodology}


In this section, we describe the methodology used to merge the weights of two pre-trained neural network models using a genetic algorithm. We call our methodology, MeGA. Genetic algorithms provide a robust optimization framework by mimicking natural selection, making them suitable for complex tasks like weight merging~\cite{Mitchell1998AnIntroduction}.

Our approach aligns with the lottery ticket hypothesis, which suggests that neural networks contain critical weights (winning tickets) necessary for maintaining and enhancing performance~\cite{frankle2019lottery}. In our MeGA algorithm, child models inherit and preserve these beneficial weight configurations from their parent models. By iteratively selecting the best-performing individuals as parents, the algorithm ensures that critical weights are retained and combined, allowing child models to evolve and enhance overall performance. This method enables us to effectively navigate the weight space, merging multiple pre-trained models into a single, superior model that leverages the strengths of each original network.

\subsection{Genetic Algorithm Framework}
\begin{figure}
\centering
\includegraphics[width=1.0\columnwidth]{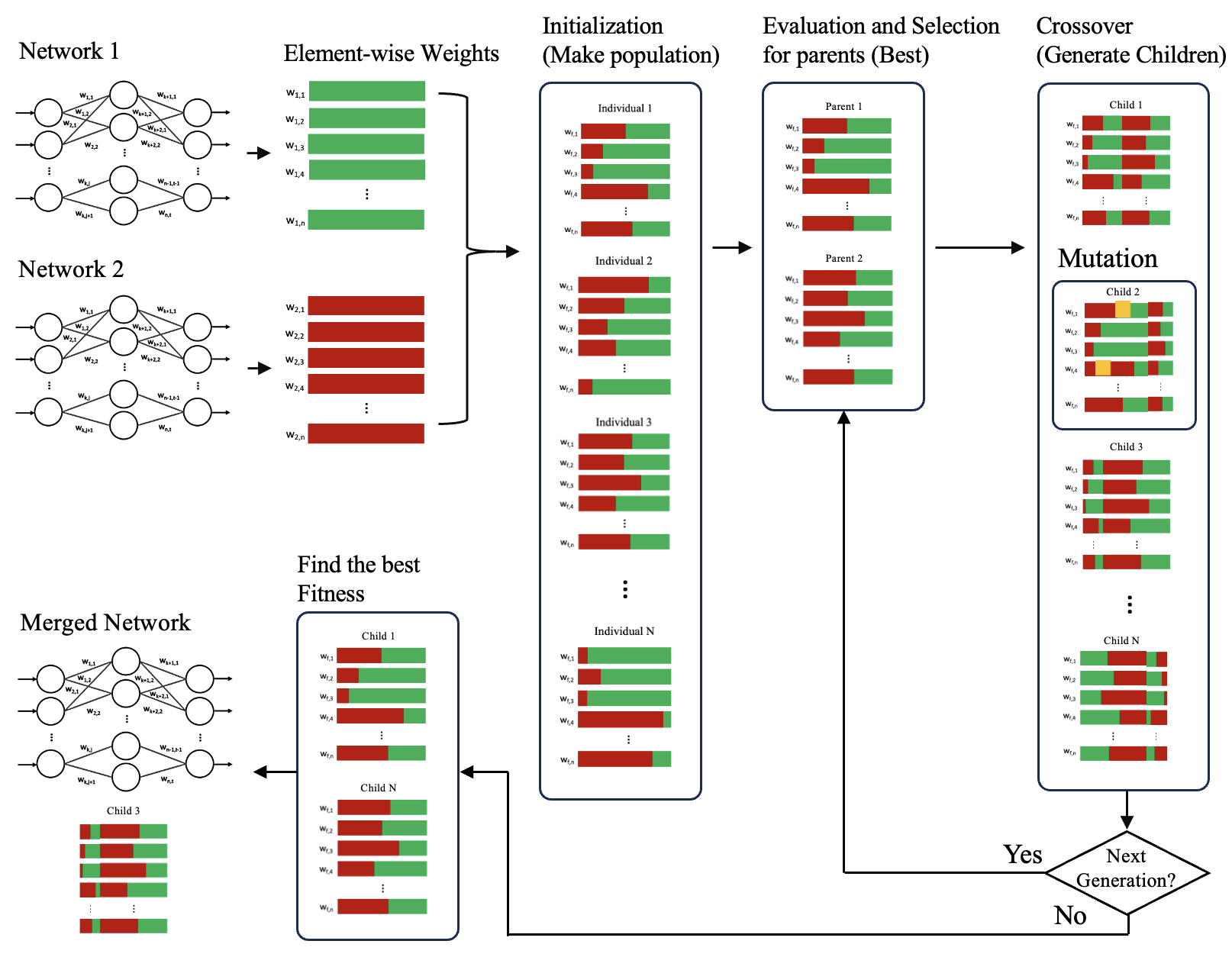}
\caption{MeGA: The process of merging two neural networks using a genetic algorithm.}
\label{fig:weights}
\end{figure}

The process of merging two neural networks using a genetic algorithm is illustrated in Figure~\ref{fig:weights}. The initial population is created by combining the element-wise weights from two pre-trained networks. Each individual in the population represents a potential solution with a unique combination of weights. The genetic algorithm iteratively selects the best-performing individuals as parents based on their evaluation, performs crossover to generate new children, and applies mutation to introduce variability. This process continues during several generations, resulting in a single merged network that combines the strengths of both original networks.

Genetic algorithms operate on a population of potential solutions, iteratively improving them through the processes of selection, crossover, and mutation~\cite{Mitchell1998AnIntroduction}. Each individual in the population represents a candidate solution, and its fitness is evaluated based on a predefined objective function. The algorithm proceeds by selecting individuals with higher fitness to produce offspring, combining their traits through crossover, and introducing random variations through mutation. Over successive generations, the population evolves towards better solutions~\cite{Mitchell1998AnIntroduction}.

\begin{algorithm}
\caption{Genetic Algorithm for Weight Merging (Element-wise)}
\begin{algorithmic}[1]
\REQUIRE \( N \): Population size
\REQUIRE \( G \): Number of generations
\REQUIRE \( K \): Number of parents for tournament selection
\REQUIRE \( p_{\text{mut}} \): Mutation probability
\REQUIRE \( \sigma \): Standard deviation for Gaussian noise in mutation
\REQUIRE \( \boldsymbol{\theta}_1 \): Weights of the first pre-trained model
\REQUIRE \( \boldsymbol{\theta}_2 \): Weights of the second pre-trained model
\ENSURE Optimal weights \( \boldsymbol{\theta}^* \)

\STATE \textbf{Initialization}: 
\STATE Initialize population \( \mathcal{P} \) with \( N \) individuals where each individual \( \boldsymbol{\theta}_i \) is a linear combination of \( \boldsymbol{\theta}_1 \) and \( \boldsymbol{\theta}_2 \)
\FOR{each individual \( n \) in \( \mathcal{P} \)}
    \STATE \( \alpha \sim \text{Uniform}(0, 1) \)
    \STATE \( \theta_{i,j,k} = \alpha \theta_{1,j,k} + (1 - \alpha) \theta_{2,j,k} \quad \forall j, k \)
\ENDFOR

\STATE \textbf{Genetic Algorithm Execution}:
\FOR{generation = 1 to \( G \)}
    \STATE \textbf{Fitness Evaluation}: Evaluate fitness \( F(\boldsymbol{\theta}_n) \) for each individual in \( \mathcal{P} \)
    \FOR{each individual \( i \) in \( \mathcal{P} \)}
        \STATE \( F(\boldsymbol{\theta}_i) = \text{Accuracy}(f(\mathbf{X}_{\text{val}}; \boldsymbol{\theta}_i), \mathbf{y}_{\text{val}}) \)
    \ENDFOR

    \STATE \textbf{Selection}: Select \( K \) parents using tournament selection
    \FOR{each selection round from 1 to \( K \)}
        \STATE Randomly select \( t \) individuals to form tournament set \( \mathcal{T} \)
        \STATE \( \mathcal{T} = \{ \boldsymbol{\theta}_{1}, \boldsymbol{\theta}_{2}, \ldots, \boldsymbol{\theta}_{t} \} \)
        \STATE Evaluate the fitness \( F(\boldsymbol{\theta}) \) of each individual in the tournament set
        \STATE Select the individual with the highest fitness as a parent: \( \boldsymbol{\theta}_{\text{parent}} = \arg\max_{\boldsymbol{\theta} \in \mathcal{T}} F(\boldsymbol{\theta}) \)
        \STATE Add \( \boldsymbol{\theta}_{\text{parent}} \) to the selected parents set
    \ENDFOR

    \STATE \textbf{Crossover}: Generate offspring through element-wise crossover of selected parents
    \FOR{each pair of parents \( (\boldsymbol{\theta}_{\text{parent1}}, \boldsymbol{\theta}_{\text{parent2}}) \) in selected parents}
        \STATE \( \alpha \sim \text{Uniform}(0, 1) \)
        \STATE \( \theta_{\text{child},j,k} = \alpha \theta_{\text{parent1},j,k} + (1 - \alpha) \theta_{\text{parent2},j,k} \quad \forall j,k \)
        \STATE Add \( \boldsymbol{\theta}_{\text{child}} \) to the offspring set
    \ENDFOR

    \STATE \textbf{Mutation}: Apply mutation to offspring
    \FOR{each individual \( \boldsymbol{\theta}_{\text{child}} \) in offspring}
        \FOR{each weight element \( \theta_{j,k} \) in \( \boldsymbol{\theta}_{\text{child}} \)}
            \IF{random number \( < p_{\text{mut}} \)}
                \STATE \( \theta_{j,k} = \theta_{j,k} + \epsilon \), where \( \epsilon \sim \mathcal{N}(0, \sigma^2) \)
            \ENDIF
        \ENDFOR
    \ENDFOR

    \STATE \textbf{Population Update}: Form new population \( \mathcal{P}' \) with the best individual from \( \mathcal{P} \) (elitism) and newly generated offspring
    \STATE Keep the best individual from \( \mathcal{P} \) based on fitness
    \STATE \( \mathcal{P} \leftarrow \mathcal{P}' \)

    \STATE Update the best individual \( \boldsymbol{\theta}^* \) if a better fitness is found in the current generation
\ENDFOR

\STATE \textbf{Output}: Best individual \( \boldsymbol{\theta}^* \) with the highest fitness

\end{algorithmic}
\end{algorithm}

\subsubsection{Preliminaries.}
Let \( \mathbf{X} \in \mathbb{R}^{n \times d} \) be the training data, where \( n \) is the number of samples and \( d \) is the dimensionality of each sample. The corresponding labels are denoted by \( \mathbf{y} \in \mathbb{R}^n \). We define two pre-trained neural networks, \( f_1(\mathbf{X}; \boldsymbol{\theta}_1) \) and \( f_2(\mathbf{X}; \boldsymbol{\theta}_2) \), where \( \boldsymbol{\theta}_1 \) and \( \boldsymbol{\theta}_2 \) represent the weights of the respective models. Our objective is to merge these weights to create a new model \( f(\mathbf{X}; \boldsymbol{\theta}) \) that achieves superior performance.

\subsubsection{Algorithm Execution.}
The genetic algorithm iterates through \( G \) generations, performing selection, crossover, and mutation at each step. The best individual \( \boldsymbol{\theta}^* \) with the highest fitness across all generations is selected as the final set of weights for the fused model. The overall procedure is summarized as follows:

\begin{enumerate}
    \item \textbf{Initialization}: Create an initial population \( \mathcal{P} \) of \( N \) individuals.
    \item \textbf{Fitness Evaluation}: Evaluate the fitness \( F(\boldsymbol{\theta}) \) for each individual in \( \mathcal{P} \).
    \item \textbf{Selection}: Select \( K \) parents from \( \mathcal{P} \) using tournament selection.
    \item \textbf{Crossover}: Generate offspring through crossover of selected parents.
    \item \textbf{Mutation}: Apply mutation to the offspring.
    \item \textbf{Population Update}: Form the new population \( \mathcal{P}' \) with the best individual from \( \mathcal{P} \) (elitism) and the newly generated offspring.
    \item \textbf{Iteration}: Repeat steps 2-6 for \( G \) generations.
\end{enumerate}




\subsubsection{Initialization}
The initial population \( \mathcal{P} \) consists of \( N \) individuals, where each individual represents a potential solution in the form of a set of weights \( \boldsymbol{\theta} \). Each individual is initialized to ensure diversity, which is crucial for the effectiveness of the genetic algorithm.

The weights of each individual \( \boldsymbol{\theta}_i \) are initialized as an element-wise linear combination of the weights from two pre-trained models, \( \boldsymbol{\theta}_1 \) and \( \boldsymbol{\theta}_2 \):
\[
\theta_{i,j,k} = \alpha \theta_{1,j,k} + (1 - \alpha) \theta_{2,j,k}, \quad \alpha \sim \text{Uniform}(0, 1)
\]
Here, \( \alpha \) is a random scalar drawn from a uniform distribution between 0 and 1. This ensures that each individual's weights are a unique blend of the two parent models, with the combination applied element-wise to preserve the detailed characteristics of both models. 

\subsubsection{Fitness Evaluation}
The fitness of each individual in the population is assessed based on the validation accuracy on a hold-out validation set \( (\mathbf{X}_{\text{val}}, \mathbf{y}_{\text{val}}) \). For an individual with weights \( \boldsymbol{\theta}_i \), the fitness function \( F(\boldsymbol{\theta}_i) \) is defined as follows:
\[
F(\boldsymbol{\theta}_i) = \text{Accuracy}(f(\mathbf{X}_{\text{val}}; \boldsymbol{\theta}_i), \mathbf{y}_{\text{val}})
\]
This function measures how accurately the model with weights \( \boldsymbol{\theta}_i \) predicts the labels of the validation set. The model \( f(\mathbf{X}; \boldsymbol{\theta}_i) \) is evaluated, and the accuracy is computed as the proportion of correctly classified samples in \( \mathbf{X}_{\text{val}} \).

Mathematically, the accuracy is given by:
\[
\text{Accuracy}(f(\mathbf{X}; \boldsymbol{\theta}), \mathbf{y}) = \frac{1}{n_{\text{val}}} \sum_{i=1}^{n_{\text{val}}} \mathbb{I} \left( \arg\max_j f_j(\mathbf{X}_i; \boldsymbol{\theta}) = y_i \right)
\]
where \( n_{\text{val}} \) is the number of validation samples, \( f_j(\mathbf{X}_i; \boldsymbol{\theta}) \) is the predicted probability for class \( j \) for the \( i \)-th validation sample, and \( \mathbb{I} \) is the indicator function that equals 1 if the predicted class matches the true label \( y_i \), and 0 otherwise.

\subsubsection{Selection}
Tournament selection is employed to choose \( K \) parents for the next generation. This method ensures that individuals with higher fitness are more likely to be selected, promoting desirable traits in subsequent generations. The selection process involves several steps.

First, we randomly select \( t \) individuals from the current population \( \mathcal{P} \) to form a tournament set \( \mathcal{T} \):
\[
\mathcal{T} = \{ \boldsymbol{\theta}_{i_1}, \boldsymbol{\theta}_{i_2}, \ldots, \boldsymbol{\theta}_{i_t} \}, \quad \mathcal{T} \subseteq \mathcal{P}
\]
Next, the fitness \( F(\boldsymbol{\theta}) \) of each individual \( \boldsymbol{\theta} \) in the tournament set \( \mathcal{T} \) is evaluated using the previously defined fitness function. After evaluating the fitness of all individuals in the tournament set, the individual with the highest fitness is selected as a parent:
\[
\boldsymbol{\theta}_{\text{parent}} = \arg\max_{\boldsymbol{\theta} \in \mathcal{T}} F(\boldsymbol{\theta})
\]
This process is repeated until \( K \) parents are selected for crossover.

\subsubsection{Crossover}
Crossover is performed on pairs of selected parents to generate offspring. This process combines the genetic material (weights) of two parents to produce a new individual (offspring), thereby promoting genetic diversity. Given two parents \( \boldsymbol{\theta}_a \) and \( \boldsymbol{\theta}_b \), an offspring \( \boldsymbol{\theta}_{\text{child}} \) is created by taking a weighted combination of the parents' weights. The crossover is performed element-wise for each weight:
\[
\theta_{\text{child}, j,k} = \beta \theta_{a, j,k} + (1 - \beta) \theta_{b, j,k}, \quad \beta \sim \text{Uniform}(0, 1)
\]

\subsubsection{Mutation}
Mutation introduces variability into the population by perturbing the weights of the offspring. For each weight \( \theta_{j,k} \) in the offspring \( \boldsymbol{\theta}_{\text{child}} \), mutation is applied with a probability \( p_{\text{mut}} \):
\[
\theta_{child, j,k} \leftarrow \theta_{child, j,k} + \epsilon, \quad \epsilon \sim \mathcal{N}(0, \sigma^2)
\]
Here, \( \epsilon \) is a random variable drawn from a normal distribution with mean 0 and variance \( \sigma^2 \). The mutation rate \( p_{\text{mut}} \) controls the likelihood of each weight being mutated. This mutation process ensures that the population maintains genetic diversity by introducing new genetic material.


\section{Experimental Results}
\subsection{Experimental Settings}
We utilized the CIFAR-10 dataset~\cite{Krizhevsky2009Learning}. The CIFAR-10 training dataset was split into 45,000 training and 5,000 validation sets. The validation set was used during the model validation and merging process. We use CNN models: ResNet~\cite{He2016ResNet}, Xception~\cite{chollet2017xception}, and DenseNet~\cite{huang2017densely} without pre-trained weights and without data augmentation. These models were trained using a batch size of 256 over 50 epochs with the Adam optimizer~\cite{Kingma2014Adam}, using a learning rate of 0.01.

Each model followed the same architecture and hyperparameters to ensure consistency and a fair comparison. After training, we applied our MeGA approach to merge the weights of these models into a single set of weights. The performance of the merged model was then evaluated and compared against the individual models. For all experiments, we used the following MeGA hyperparameters: a population size of 20, 20 generations, 4 parents per generation, and a mutation rate of 0.02.

\subsection{CIFAR-10 Results}
The results of the image classification experiments on the CIFAR-10 dataset~\cite{Krizhevsky2009Learning} demonstrate the effectiveness of our MeGA. 

\setlength{\tabcolsep}{2pt}
\begin{table}[htbp]
\centering
\caption{Performance Comparison of Individual and Merged CNN Models based on MeGA with CIFAR-10 Dataset}
\begin{adjustbox}{width=1\textwidth}
\begin{tabular}{@{}l|lllclc@{}}
\toprule
\toprule
Datasets & Model (\# of Params.) & & Baseline. & Baseline Test Acc. & Merging Method & Final Test Acc.  \\
\# of classes && & (Random Seed) & (Single Model) & & (Merged Model)  \\

\toprule
&\multirow{2}{*}{ResNet 56 (0.86M)~\cite{He2016ResNet}} & \ \ \ & Model 1 (56) & 0.809 & Weight Average &  0.010\\
                                    &&& Model 2 (57) & 0.793 &MeGA (Ours) &  0.822 \\
                  \cline{2-7}              
&\multirow{2}{*}{ResNet 110 (1.78M)~\cite{He2016ResNet}} && Model 1 (56) & 0.801 &Weight Average &  0.010 \\
                                     &&& Model 2 (57) & 0.814 &MeGA (Ours) &  0.816 \\
                                     
                   \cline{2-7}                  
&\multirow{2}{*}{ResNet 152 (3.51M)~\cite{He2016ResNet}} && Model 1 (56) & 0.783 &Weight Average &  0.010 \\
CIFAR-10~\cite{Krizhevsky2009Learning}   &&& Model 2 (57) & 0.781 &MeGA (Ours) & 0.819 \\

                   \cline{2-7}                
(10) &\multirow{2}{*}{Xception (22.96M)~\cite{He2016ResNet}} && Model 1 (56) & 0.716 &Weight Average & 0.010 \\
                                    &&& Model 2 (57) & 0.730 &MeGA (Ours) & 0.754 \\
                   \cline{2-7}                 
&\multirow{2}{*}{DenseNet 121 (7.04M)~\cite{He2016ResNet}} && Model 1 (56) & 0.719 &Weight Average & 0.010 \\
                                    &&& Model 2 (57) & 0.728 &MeGA (Ours) & 0.742 \\
                    \cline{2-7}              
&\multirow{2}{*}{DenseNet 169 (12.65M)~\cite{He2016ResNet}} && Model 1 (56) & 0.735 & Weight Average &  0.010 \\
                                    &&& Model 2 (57) & 0.731 & MeGA (Ours) & 0.753 \\
\bottomrule
\bottomrule
\end{tabular}
\label{tab:cnn}
\end{adjustbox}
\end{table}

The MeGA approach led to significant improvements in test accuracies across various models as shown in the Table~\ref{tab:cnn}. For example, the ResNet 56 model's accuracy improved from 0.801 to 0.822, and the ResNet 110 model reached 0.824. 

\begin{figure}
\centering
\includegraphics[width=1.0\columnwidth]{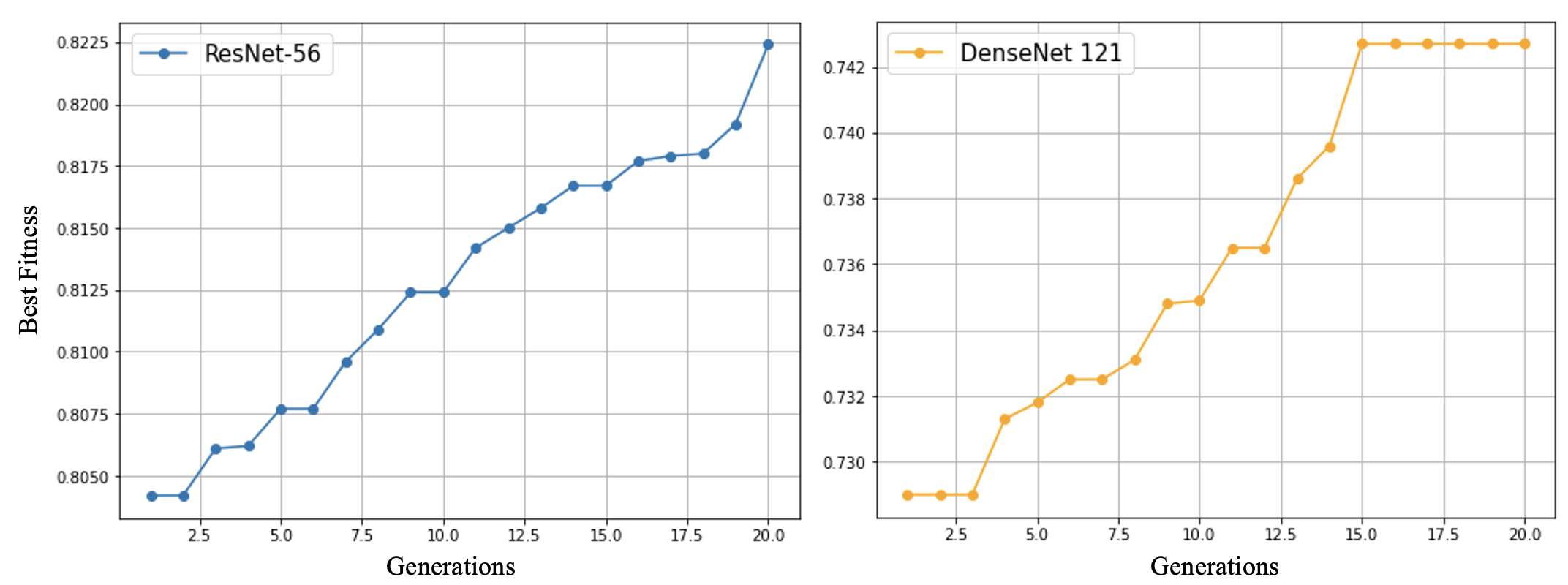}
\caption{MeGA Fitness Progression (Test Accuracy) for ResNet-56 and DenseNet 121 models over 20 generations.}
\label{fig:MeGA_fitness_progression}
\end{figure}

The Xception model's accuracy increased from 0.723 to 0.754, demonstrating the effectiveness of MeGA in optimizing complex architectures. Similarly, the DenseNet 121 model's accuracy rose from 0.724 to 0.742, and the DenseNet 169 model improved from 0.733 to 0.753. Overall, the MeGA approach consistently outperformed individual models, highlighting its potential for enhancing neural network performance through genetic algorithm-based weight merging.

Figure~\ref{fig:MeGA_fitness_progression} shows the progression of the best fitness values over 20 generations for the ResNet-56 and DenseNet 121 models. For the ResNet-56 model, the best fitness improved steadily from 0.8042 in the first generation to 0.8224 in the twentieth generation. Similarly, the DenseNet 121 model saw an improvement in best fitness from 0.7290 to 0.7427 over the same period. These plots illustrate the genetic algorithm's capability to effectively navigate the weight space and optimize combinations to enhance model performance.

The reason why the weight averaging method resulted in poorer performance is that the networks were initialized differently and trained independently. This led to discrepancies in the learned weights, making simple averaging ineffective. This highlights the importance of considering the initialization and training processes when merging models.

\subsection{Extended Experiments for Multi-Models}
\begin{figure}
\centering
\includegraphics[width=1.0\columnwidth]{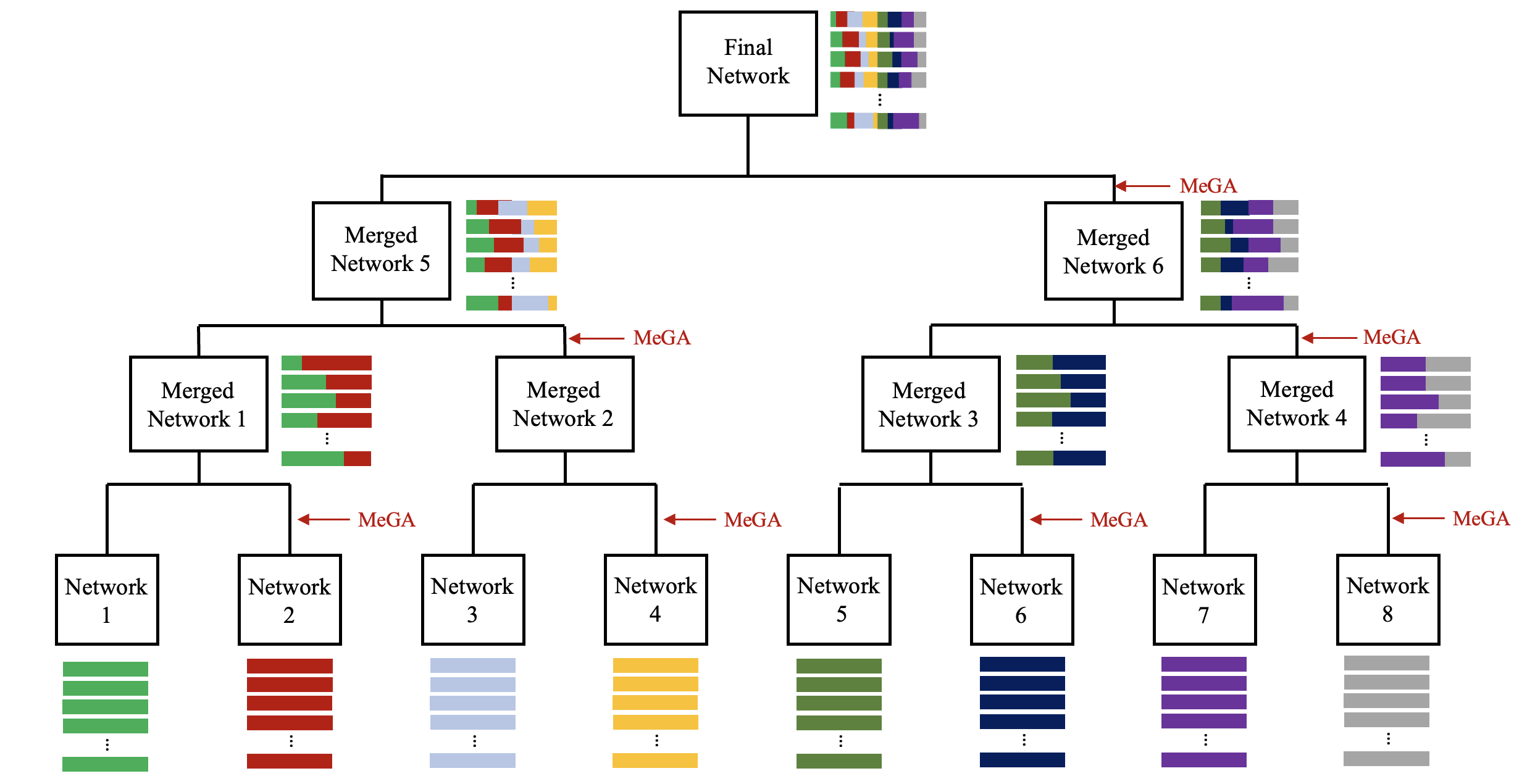}
\caption{Hierarchical merging process of eight models into a single final model.}
\label{fig:merge}
\end{figure}

In this section, we illustrate the application of our genetic algorithm-based weight merging method to combine multiple neural network models as depicted in Figure~\ref{fig:merge}. The hierarchical merging process starts by training eight models on the CIFAR-10 dataset~\cite{He2016ResNet, Krizhevsky2009Learning}. These models are paired and merged using a genetic algorithm, resulting in four intermediate models. These intermediate models are further merged into two higher-level models, which are finally merged into a single, robust model. 

\setlength{\tabcolsep}{3pt}
\begin{table}[htbp]
\centering
\caption{Performance Comparison of 8 Individual and Merged ResNet Models based on MeGA with CIFAR-10 Dataset}
\begin{adjustbox}{width=1\textwidth}
\begin{tabular}{@{}l|lllclc@{}}
\toprule
\toprule
Dataset &Model (\# of Params) & & Baseline. & Baseline Test Acc. & Merging Method & Final Test Acc.  \\
\# of classes &&& (Random Seed) & (Single Model) & &(Merged Model)\\
\midrule
&\multirow{8}{*}{ResNet 56 (0.86M)~\cite{He2016ResNet}} & & Model 1 (44) & 0.804 & \multirow{4}{*}{Weight Average} & \multirow{4}{*}{0.010} \\
&& & Model 2 (45) & 0.747 & \\
&& & Model 3 (46) & 0.788 & \\
&& & Model 4 (47) & 0.774 & \\
&& & Model 5 (48) & 0.806 & \multirow{4}{*}{MeGA (Ours)} & \multirow{4}{*}{0.822} \\
&& & Model 6 (49) & 0.786 & \\
&& & Model 7 (50) & 0.749 & \\
&& & Model 8 (51) & 0.787 & \\
\cline{2-7}   
&\multirow{8}{*}{ResNet 110 (1.78M)~\cite{He2016ResNet}} & & Model 1 (44) & 0.782 &\multirow{4}{*}{Weight Average} & \multirow{4}{*}{0.010} \\
&& & Model 2 (45) & 0.814 & \\
&& & Model 3 (46) & 0.744 & \\
CIFAR-10~\cite{Krizhevsky2009Learning}&& & Model 4 (47) & 0.800 & \\
(10)&& & Model 5 (48) & 0.801 & \multirow{4}{*}{MeGA (Ours)} & \multirow{4}{*}{0.824}\\
&& & Model 6 (49) & 0.787 & \\
&& & Model 7 (50) & 0.781 & \\
&& & Model 8 (51) & 0.758 & \\
\cline{2-7}   
&\multirow{8}{*}{ResNet 152 (3.51M)~\cite{He2016ResNet}} & & Model 1 (44) & 0.775 &\multirow{4}{*}{Weight Average}& \multirow{4}{*}{0.010} \\
&& & Model 2 (45) & 0.783 & \\
&& & Model 3 (46) & 0.781 & \\
&& & Model 4 (47) & 0.747 & \\
&& & Model 5 (48) & 0.771 & \multirow{4}{*}{MeGA (Ours)} & \multirow{4}{*}{0.815}\\
&& & Model 6 (49) & 0.767 & \\
&& & Model 7 (50) & 0.772 & \\
&& & Model 8 (51) & 0.797 & \\

\bottomrule
\bottomrule
\end{tabular}
\end{adjustbox}
\label{tab:cnn}
\end{table}

This process ensures the final model combines the strengths of all eight original models. The results in Table~\ref{tab:cnn} demonstrate the effectiveness of our genetic algorithm-based weight merging method. For ResNet 56 models, baseline test accuracies ranged from 0.747 to 0.806, improving to 0.822 after merging. ResNet 110 models had baseline accuracies between 0.744 and 0.814, with the merged model achieving 0.824. For ResNet 152 models, baseline accuracies ranged from 0.747 to 0.783, and the merged model reached 0.815. Despite the larger complexity of ResNet 152, our method effectively enhanced model performance.

These results highlight the potential of our approach to enhance neural network performance. The consistent improvement in test accuracy across different architectures (ResNet 56, ResNet 110, and ResNet 152) demonstrates the versatility and scalability of our method. The hierarchical merging process ensures the final model benefits from the strengths of individual models.


\section{Discussion}
The application of genetic algorithm-based weight merging to combine multiple neural network models presents several significant advantages. This method is particularly beneficial for leveraging the strengths of various independently trained models, leading to enhanced overall performance. Belows are several significant advantages:

\begin{itemize}
    \item \noindent\textbf{Capturing Complementary Features:} Merging multiple neural networks captures complementary features learned by individual models. Each model specializes in recognizing different patterns within the data, and combining their weights results in improved accuracy and robustness.
    \item \noindent\textbf{Support for Distributed Environments:} This method supports efficient and scalable training in distributed environments. By enabling the merging of models trained across multiple GPUs or devices, the approach facilitates distributed training, which is valuable in cloud-based systems or edge computing environments.
    \item \noindent\textbf{Reduced Inference Resource Usage:} Using a single, high-performance merged model reduces inference resource usage compared to using multiple models. This is particularly beneficial in environments where computational resources are limited or where efficiency is critical, such as mobile or embedded systems.
\end{itemize}

The genetic algorithm-based weight merging technique significantly enhances the performance of neural networks by effectively combining multiple pre-trained models. This method improves accuracy and generalization while offering practical benefits in terms of training efficiency, scalability, and resource usage. The hierarchical merging approach underscores the potential of genetic algorithms in optimizing neural network weights, providing a robust and efficient tool for modern deep learning applications.

\section{Conclusion}
In this paper, we introduced a genetic algorithm-based method for merging the weights of multiple pre-trained neural networks. This approach demonstrated significant improvements in model performance and robustness by effectively combining the strengths of individual models. Our experiments on the CIFAR-10 dataset confirmed that the hierarchical merging process of eight ResNet56 models results in a final model with superior accuracy and generalization compared to traditional methods like weight averaging and ensemble techniques.

The genetic algorithm's ability to optimize weight combinations through selection, crossover, and mutation allows for the creation of a more effective merged model without the need for additional training or architectural changes. This method also supports scalable and efficient training in distributed environments, making it a practical solution for various deep learning applications.

Overall, the genetic algorithm-based weight merging technique offers a powerful tool for enhancing neural network performance, providing a robust and efficient framework for integrating multiple pre-trained models. This work highlights the potential of genetic algorithms in neural network optimization, paving the way for further research and applications in artificial intelligence and machine learning.

%
%
%
%
%
%
\bibliographystyle{splncs04}
%
\bibliography{main}

\end{document}